\documentclass[final,5p,times,twocolumn]{elsarticle} 

\usepackage{amsmath}
\usepackage{amssymb}
\usepackage{graphicx}
\usepackage{tabularx}
\usepackage{booktabs}
\usepackage{algorithm}
\usepackage{algpseudocode}

\usepackage{hyperref}

\begin{document}

\begin{frontmatter}

\title{Self-Directed Task Identification}

\author[1]{Timothy Gould}
\author[1]{Sidike Paheding}

\affiliation[1]{organization={Fairfield University},
                city={Fairfield},
                state={CT},
                country={USA}}

\begin{abstract}
In this work, we present a novel machine learning framework called Self-Directed Task Identification (SDTI), which enables models to autonomously identify the correct target variable for each dataset in a zero-shot setting without pre-training. SDTI is a minimal, interpretable framework demonstrating the feasibility of repurposing core machine learning concepts for a novel task structure. To our knowledge, no existing architectures have demonstrated this ability. Traditional approaches lack this capability, leaving data annotation as a time-consuming process that relies heavily on human effort. Using only standard neural network components, we show that SDTI can be achieved through appropriate problem formulation and architectural design. We evaluate the proposed framework on a range of benchmark tasks and demonstrate its effectiveness in reliably identifying the ground truth out of a set of potential target variables. SDTI outperformed baseline architectures by 14\% in F1 score on synthetic task identification benchmarks. These proof-of-concept experiments highlight the future potential of SDTI to reduce dependence on manual annotation and to enhance the scalability of autonomous learning systems in real-world applications.
\end{abstract}


\end{frontmatter}


\section{Introduction}
\label{sec:intro}
Deep learning architectures remain at the forefront of machine learning research due to their effectiveness, scalability, and widespread accessibility. Their success reflects the creativity and diligence of researchers who continuously design novel models capable of performing increasingly complex tasks. This innovation is possible in part because deep learning models must be differentiable with respect to a cost function and require access to ground truth labels in supervised learning frameworks \cite{b7}, leaving much room for creativity in model architecture design. As a result, deep learning has led to remarkable advances in natural language processing, computer vision, and robotics \cite{alom2019state}.

Despite this progress, the development and deployment of deep learning models continue to rely heavily on human expertise. Researchers must manually design training pipelines and curate annotated datasets \cite{b2,b3}, both of which are labor-intensive processes that struggle to scale with the exponential growth of available data. Although fields such as automated machine learning (AutoML), neural architecture search (NAS), and meta-learning have made significant strides in automating portions of the model development lifecycle \cite{b2,b3,b4,b5}, they do not yet enable models to autonomously determine the correct target variables or ground truth without human supervision. Existing work in this area remains largely theoretical, lacking concrete, empirically tested model architectures capable of performing this exact function \cite{b1}.

In this paper, we introduce a novel model architecture designed to enable self-directed task identification (SDTI), which allows a machine learning model to determine a task based on a given dataset without external guidance. SDTI refers to a model’s capacity to infer the correct dataset-target variable pairings when presented with multiple datasets and their associated labels, without prior knowledge of which dataset corresponds to which target variable. Conceptually, this task relates to domains such as dataset alignment, multi-task zero-shot transfer, unsupervised label matching, and unsupervised manifold alignment, all of which aim to reconcile multiple data sources without supervision. However, SDTI goes beyond aligning distributions or shared representations—it autonomously identifies the correct target variable for each dataset by leveraging the inherent structure and complexity of the data itself as an implicit supervisory signal. SDTI can be more broadly related to unsupervised manifold alignment, where models seek to align data from different domains without labels. However, unlike traditional manifold alignment—which focuses on mapping latent spaces—SDTI leverages manifold geometry itself without additional transformations to infer the correct dataset–label pairings, using complexity as an implicit supervisory signal rather than alignment objectives.

Our proposed SDTI model achieves this using simple, well-established components arranged in a unique architecture tailored for task identification. The foundation is a basic artificial neural network (ANN) that applies a weighted sum followed by a nonlinear activation function. Embedded within this framework is a novel SDTI layer, which consists of a single-neuron ANN for each dataset and target variable combination within the input corpus. To ensure efficiency, the SDTI layer incorporates two key optimizations. First, a vectorized implementation enables all single-neuron ANNs to execute in parallel. Second, each ANN does not need to learn a production-ready mapping; it only needs to distinguish its associated dataset from incorrect dataset–target variable pairings, leveraging the resistance of sub-optimal  manifolds to optimization as an implicit learning signal. This design allows SDTI to self-direct its task inference, making it a distinct and scalable solution in scenarios where traditional automated target variable identification methods would fail.

The SDTI model proceeds over $E$ iterations (Alg. \ref{alg:sdti_workflow} Line. \ref{line:maintrainingloop}). During each iteration, the SDTI layer executes $S$ training passes nested within $E$ iterations (Alg. \ref{alg:sdti_workflow} Line. \ref{line:stdiloop}). After these passes, it computes the cost for every dataset and target variable combination (Alg. \ref{alg:sdti_workflow} Line. \ref{line:cost}). The model then selects the combination with the lowest cost as the predicted target variable for a given dataset. Across $E$ iterations, the model records these predictions, and the final output is determined by selecting the combination that achieved the lowest cost after $S$ iterations within the SDTI layer most frequently (Alg. \ref{alg:sdti_workflow} Line. \ref{line:prediction}).

At present, SDTI is a minimal and interpretable framework demonstrating feasibility at a foundation level. However, the primary future application of the SDTI model is the automation of data annotation, a key bottleneck in scalable supervised learning. By removing the dependency on human-curated pipelines, SDTI has the potential to significantly reduce the time and labor required for model training. Additional use cases include optimizing large language model (LLM) fine-tuning workflows, where automated label assignment could improve prompt–completion pairing. More broadly, SDTI may enable continuous learning systems that autonomously adapt to new data, making it a promising tool for any domain requiring dynamic, scalable model training without ongoing human intervention. We will provide a GitHub repository to facilitate further research. 

\subsection{Related Works}
In this section, several works related to the study of SDTI will be analyzed in relation to the methods proposed above. Recent advances in self-supervised learning (e.g., SimCLR\cite{b18}, BYOL\cite{b19}) and zero-shot representation learning (e.g., CLIP\cite{b20}, CoOp\cite{b21}) demonstrate that high-level semantic structure can emerge from data through alignment of internal representations rather than explicit supervision. The Self-Directed Task Identification (SDTI) framework extends this principle by enabling an agent to autonomously infer dataset-target variable pairings by iteratively leveraging the values from the cost function as a proxy for manifold structure, without reliance on external pretraining or labeled correspondences. Conceptually, SDTI can be viewed as bridging unsupervised manifold alignment—which aligns geometric structures across datasets—with the data-centric and AutoML perspective, where the discovery of coherent internal structure becomes the optimization target itself. This situates SDTI within the broader effort to generalize self-supervised and zero-shot paradigms from representation alignment across data to task discovery by leveraging the the model’s own latent space, advancing toward systems capable of self-organizing task definitions directly from structure.

Automated Machine Learning (AutoML) refers to the process of automating key steps within the machine learning pipeline. Tasks such as feature engineering, model selection, and hyperparameter-tuning all require an understanding of statistics and machine learning concepts. The purpose of AutoML is to empower non-experts to develop and deploy machine learning models. Studies such as \cite{b2}–\cite{b3} have examined the concept of AutoML, highlighted its progress in recent years, and discussed potential future trends. While AutoML is an impactful area of study, human judgment in defining the learning objective remains critical to its successful implementation. The proposed SDTI model aims to extend the scope of AutoML research by incorporating SDTI. 

Central to this paper is the concept of manifold complexity. Previous attempts sought to improve the performance of machine learning models by way of semi-supervised learning. In exploring this concept, they discussed that the addition of unlabelled data points reduces the complexity of the underlying function represented by the manifold by way of reducing the sparsity of the dataset \cite{b16}. Sparsity in the data can lead to a complicated manifold, as the small amount of data may not adequately reflect the semantic relationship between the data and ground truth, making the manifold more difficult for the algorithm to navigate. Rather than trying to eliminate manifold complexity, the SDTI model leverages it to properly identify which target variable corresponds to each dataset. 
\section{Experimental Design}
We evaluate our method using 17 benchmark datasets commonly employed in academic research and publicly available repositories. These datasets are selected to provide a diverse and representative testbed for zero-shot prediction. The simplistic benchmarks used were intended to mirror the capabilities of SDTI as SDTI is a minimal, interpretable proof-of-concept demonstrating the feasibility of repurposing core machine learning concepts for a novel task structure.

Each dataset, denoted as $D$ herein, is independently normalized using z-score normalization, computed only from its training subset to prevent data leakage. Importantly, each dataset contains a single ground-truth or target variable, which serves as the target for prediction.

The SDTI model is evaluated within a zero-shot learning framework, where no prior information is provided about the correct target variable assignments or dataset identities. Instead of training on pre-defined input and target variable pairs, the model must infer the correct alignment across datasets autonomously. Internally, the model explores various candidate data and target variable combinations using supervised components, but without access to task-specific guidance. This setup preserves the integrity of zero-shot evaluation while enabling systematic testing across multiple datasets.

\subsection{Datasets}
Only normalized data are used for each dataset in the corpus, with at most 2,000 samples selected from each dataset. Normalization is performed independently for each dataset using its own normalization weights, rather than applying a shared normalization scheme across the entire corpus.

To ensure consistency in input dimensions, all datasets, including CIFAR-10, Fashion-MNIST, Digit-MNIST, and SVHN, are flattened into two-dimensional vectors suitable for training with an artificial neural network (ANN) architecture. These datasets are specifically chosen because the semantic relationships within the flattened images and their corresponding labels are subtle and difficult to detect. This complexity makes them well-suited for evaluating model performance under challenging learning conditions and for stress-testing the SDTI model.

Datasets with multi-class target variables are partitioned into binary classification problems for two key reasons. First, evaluating SDTI’s ability to match datasets with their correct target variables is more rigorous when both originate from the same domain, reflecting real-world conditions where related tasks often share overlapping data spaces. Second, binary classification increases task difficulty—each ANN faces a one-in-two chance of a correct prediction, which can introduce slight noise but also increases the risk of incorrectly associating datasets with unrelated target variables, a problem less likely in multiclass setups.

\subsection{Training Process}
Each SDTI model runs for $E$ of iterations, and the SDTI layer runs each ANN for $S$ iterations within each of the SDTI models $E$ iterations throughout the training process. The vectorized nature of the SDTI layer ensured that each ANN could learn simultaneously, which significantly decreased the time to train each SDTI model. Without vectorization, each iteration takes approximately 5 minutes or 10 hours per SDTI model in our experiments. The vectorized implementation takes approximately 3 seconds per iteration in our case. The experiment is performed on the GPU of an AWS Sagemaker ml.g4dn.xlarge instance featuring one NVIDIA T4 Tensor Core GPU with 16 GB of GPU memory, 4 vCPUs, and 16 GB of system memory.

\subsection{Testing Process}
To evaluate the SDTI model, we systematically varied three parameters: the number of training records $A$, the number of model training epochs $B$, and the number of SDTI layer iterations $C$. Each parameter is iterated over a fixed discrete range as follows.
\begin{align*}
A &= \{\, 2000 - 100k \mid k \in \mathbb{Z},\ 0 \leq k \leq 19 \,\} \\
B &= \{\, 2k + 1 \mid k \in \mathbb{Z},\ 0 \leq k \leq 29 \,\} \\
C &= \{\, 30 - 5k \mid k \in \mathbb{Z},\ 0 \leq k \leq 5 \,\} \\
A \times B \times C &= \{\, (a, b, c) \mid a \in A,\ b \in B,\ c \in C \,\}
\end{align*}

The Cartesian product $A \times B \times C$ defines the full set of hyperparameter combinations tested. A total of 3,600 sampled combinations from this space were evaluated, with the SDTI model trained and tested separately on each configuration.

\subsection{Model Evaluation}
The SDTI model was evaluated primarily using the F1 score. Additional evaluation metrics included raw classification accuracy and inference time (measured in minutes) to assess the efficiency of the model when processing large sets of dataset-label combinations in parallel.

Additional benchmark tests were conducted using established algorithms to provide comparative baselines against the proposed SDTI framework. Because SDTI’s objective is to evaluate internal task structure consistency rather than to optimize cross-domain embedding alignment, we employ statistical similarity measures (Pearson correlation, mutual information, and cosine similarity) as baseline metrics. These measures directly quantify the stability and coherence of the manifolds for each dataset-target variable combination.

Unsupervised manifold techniques were not chosen as their primary function is to transform and re-embed data into new coordinate systems to achieve alignment. Likewise, models like CLIP that are capable of pairing dataset-target variable pairs in the form of image and text captions rely heavily on pretraining from external datasets \cite{b20}. SDTI's configuration is stateless and requires no pretraining for zero-shot predictions making models like CLIP an inaccurate comparison.

We conducted an ablation study to evaluate the model’s robustness under reduced information conditions. In these experiments, we systematically reduced the number of records, the number of overall training iterations, and the number of iterations within the SDTI layer. The model retained reasonable predictive performance even with limited exposure to data, demonstrating its efficiency in zero-shot inference settings.

\subsection{Limitations}
 The limitations in this section are broad on account of the fact that SDTI is a minimal, interpretable proof-of-concept demonstrating the feasibility of repurposing core machine learning concepts for a novel task structure.

The main limitation of SDTI is scalability. Although vectorization speeds up inference, memory grows rapidly with more datasets. For the 17 datasets here, all $17^2 = 289$ dataset–target combinations are processed in parallel, each requiring a set of weights. Expanding to 20 datasets increases combinations to 400, straining memory. A batching framework is needed to flexibly trade off speed and memory for larger deployments.

Currently, the SDTI model is limited to tabular data involving binary classification tasks. Given the rapid advancement of multimodal learning \cite{b6} and the relevance of multimodal integration for downstream applications such as feature engineering and virtual assistants, extending SDTI to support multimodal inputs is a critical next step. Future work will be required to adapt the SDTI architecture to operate effectively across modalities, including vision, text, and audio.

The model is also sensitive to class imbalance. If a label set is mostly zeros, the corresponding ANN may collapse the weights to near-zero outputs, falsely indicating a good fit. Mitigating this likely requires refined loss functions or normalization to handle imbalance.

In a production environment, the correct target variable for a dataset may not be in the current batch or in the entire corpus of data being examined. Currently, there is no unknown token within the model configuration which would limit the real world effectiveness of the SDTI model.

\section{Model Structure}
In this work, we explore the possibilities of SDTI by leveraging a vectorized implementation of multiple ANNs existing within the same matrices, but learning the data independently of one another. This methodology will be used in an attempt to test all available datasets and target variable combinations across 17 datasets. Ultimately, the SDTI model will determine which dataset and target variable combination produces the least amount of cost most frequently across $E$ epochs to determine which target variable belongs to each dataset. Alg \ref{alg:sdti_workflow} illustrates the SDTI Training and Prediction Workflow. 

\begin{figure*}[h]
    \centering
    \includegraphics[width=\textwidth]{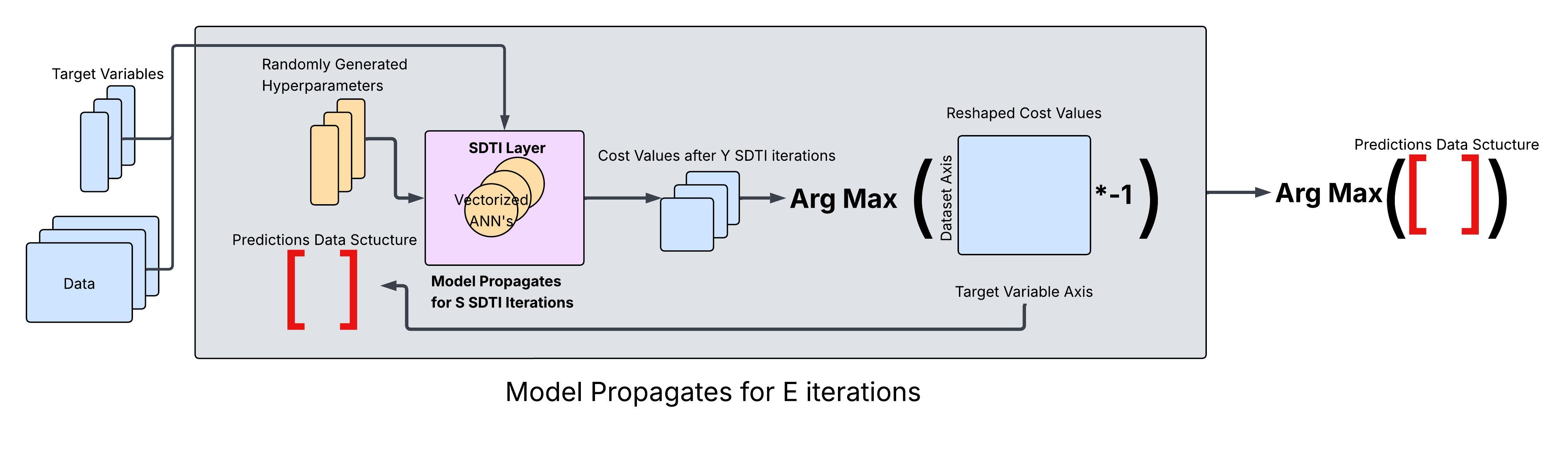}
    \caption{Visual representation of Alg \ref{alg:sdti_workflow}. Within the outer loop the model randomly creates hyperparameters to be passed into the SDTI layer for the vecotrized ANN's running simultaneously. The SDTI layer will run for $S$ iterations and return the last cost value produced. These values are then reshaped and passed into an ArgMax function. This output is then saved in a specific data structure $T$ that is passed through another ArgMax function after $E$ iterations. The final output is a vector of integers indicating predicted target variable for a given dataset.} 
    \label{fig:model_schema}
\end{figure*}

\begin{algorithm}[htbp]
\caption{SDTI Training and Prediction Workflow}
\label{alg:sdti_workflow}
\centering
\begin{minipage}{1.0\columnwidth} 
\begin{algorithmic}[1]
\State Initialize empty list $T$ to store predictions per epoch

\For{epoch $= 1 \to E$} \Comment{Main training epochs} \label{line:maintrainingloop}
    \State Sample learning rates $\{lr_i\}$ and betas $\{\beta_{1i}, \beta_{2i}\}$ \label{line:hpinit}
    \State Initialize weights $W$ and biases $b$ in SDTI Layer \label{line:winit}
    
    \For{sdti\_epoch $= 1 \to S$} \Comment{Nested SDTI epochs} \label{line:stdiloop}
        \State Perform forward propagation for 
        \Statex \quad all data/target variable combinations \label{line:forwardprop}
        \State Compute cost for each dataset/target variable combination \label{line:cost}
        \State Compute gradients of $W$ and $b$ \label{line:gradients}
        \State Update weights $W$ and biases $b$ \label{line:update}
    \EndFor
    
    \State Compute final cost $C \gets -1 \cdot C$
    \State Determine predicted label: $y \gets \arg\max(C)$
    \State Append prediction $y$ to list $T$ \label{line:append}
\EndFor

\State Compute overall prediction: $\hat{y} \gets \arg\max(\sum T)$  \label{line:prediction}
\State \Return Final prediction $\hat{y}$
\end{algorithmic}
\end{minipage}
\end{algorithm}

\newcommand{\MetadataMatrix}{Q_{c \times d \times g}}
\newcommand{\OutputOne}{O^{L1}_{c \times d \times g}}
\newcommand{\OutputOneReshapeOne}{O^{L1}_{d \times c \times g}}
\newcommand{\OutputOneReshapeTwo}{O^{L1}_{d \times c * g}}
\newcommand{\LearningRates}{O^{L2}_{[:,0]}}
\newcommand{\Betas}{O^{L2}_{[:,1:2]}}

\newcommand{\SDTIlayerSet}{\{\Omega((d,h),p) \mid d \in D, h \in H, p \in P\}}

\subsection{Hyperparameter Generation}
\textbf{}
Hyperparameters $h$ are generated randomly at each iteration of the SDTI model. Learning rates $lr$ are sampled from a log-uniform distribution (Eq.~\ref{eq:log uniform sampling}), and the parameters $\beta$ are drawn from uniform distributions ($\mathcal{U}$) as specified in (Eq.~\ref{eq:beta_sampling}). The number of unique hyperparameters generated is determined by the number of datasets ($D$) in the environment, denoted by $N_{D}$. Depending on the hyperparameter, the number of columns ($s$) may vary (Eq.~\ref{eq:hp sizes}) depending on whether or not ($lr$) or ($\beta$) are being generated within the hyperparameter tensor (Eq.~\ref{eq:hp gen}).
\begin{equation}
    lr_i = 10^{\,u_i}, \quad u_i \sim \mathcal{U}(-5, -1), \quad i = 1, \dots, N_{D}
    \label{eq:log uniform sampling}
\end{equation}
\begin{equation}
  \begin{aligned}
    (\beta_{1,i}, \beta_{2,i}) 
    &\sim \mathcal{U}(0.85,\,0.99) \times \mathcal{U}(0.98,\,0.9999)
  \end{aligned}
  \label{eq:beta_sampling}
\end{equation}

\begin{equation}
    s = \begin{cases} 
     1, & \text{if } h = lr \\
     2, & \text{if } h = \beta
    \end{cases} 
    \label{eq:hp sizes}
\end{equation}
\begin{equation}
    h^j = \begin{bmatrix}
    h^j_{1\times s}, h^j_{1\times s}, \dots, h^j_{1\times s}
    \end{bmatrix}, \quad j \in \{1,\dots,N_{D}\}
    \label{eq:hp gen}
\end{equation}

These hyperparameters are then stacked into a matrix, denoted $H$, suitable for vectorized implementation, enabling multiple ANNs to learn simultaneously—but independently within the same structure (Eq.~\ref{eq:hp stack}). Separate matrices are maintained for different types of hyperparameters, so $H$ can represent either the learning rates ($lr$) or the betas ($\beta$) in the context of the ADAM optimizer \cite{b17}.

\begin{equation}
    H = \begin{bmatrix}
        h^1, h^2, \dots , h^{N_D}
    \end{bmatrix} \in \mathbb{R}^{N_D^2 \times s}
    \label{eq:hp stack}
\end{equation}

This stacked matrix is subsequently used as part of an Adam optimizer, allowing for the simultaneous updating of $N_D^2$ ANNs within the SDTI layer.

\subsection{SDTI layer}
The SDTI layer consists of single-neuron ANNs, $\Omega$, trained on each dataset–target pair $(D, Y)$. Each dataset has hyperparameters ($H$) randomly generated per epoch but fixed during nested SDTI epochs. To prevent bottlenecks and improve gradient flow, the ANNs are implemented in a vectorized manner. Since datasets have different numbers of features, each $D_{m\times n}$ is zero-padded with $Z_{m\times (c-u)}$—where $m$ is the number of samples, $c$ the maximum number of columns, and $u$ the original feature count—so all datasets can be concatenated into a single tensor (Eq. \ref{eq:zero-padding}).

\begin{equation}
    a_{m \times n} = Concat(D_{m\times n},Z_{m \times (c-u)})
    \label{eq:zero-padding}
\end{equation}

Once datasets are dimension-matched, they are concatenated into a three-dimensional matrix $A_{N_D^2 \times m \times n}$ so each dataset–label pair can be tested simultaneously. $A$ has $N_D^2$ entries because each dataset $a^j$ is tested against multiple noisy label sets (Eq. \ref{eq:NoiseLabel Logic}), where one target variable is the ground truth and others are false labels. The tensor is organized so that identical datasets occupy neighboring indices (Eqs. \ref{eq:Data Matrix P1} and \ref{eq:Data Matrix P2}).

\begin{equation}
    a^j = \begin{bmatrix}
    a^j_{m\times n}, a^j_{m\times n},..., a^j_{m\times n}
    \end{bmatrix}, j \in \{1,...N_D\}
    \label{eq:Data Matrix P1}
\end{equation}
\begin{equation}
    A = \begin{bmatrix}
        a^1, a^2, ... , a^{N_D}
    \end{bmatrix} \in R^{N_D^2 \times m \times n}
    \label{eq:Data Matrix P2}
\end{equation}

The noisy label tensor $NL$, adopts a slightly different configuration to ensure that each combination of data and noisy labels is properly tested. The noisy label tensor $NL_{N_D^2 \times m \times 1}$ is structured so that identical sets of target variables do not share neighboring indices, as shown in (Eq. \ref{eq:NoiseLabel Matrix P1}). 
\begin{equation}
    nl = \begin{bmatrix}
    nl^1, nl^2, ... , nl^{N_D}
    \end{bmatrix}
    \label{eq:NoiseLabel Matrix P1}
\end{equation}
\begin{equation}
    NL = \begin{bmatrix}
    nl_{m\times 1}^1, nl_{m\times 1}^2, ... , nl_{m\times 1}^{N_D}
    \end{bmatrix}
    \label{eq:NoiseLabel Matrix P2}
\end{equation}

\noindent where $NL$ is the noisy label tensor. If $TL$ represents the true labels and $FL$ are false labels, then $FL$ are elements within $NL$ such that $TL, FL \in NL$. Each dataset utilized in this experiment will have different sets of $TL$ and $FL$ depending on which dataset $a^j$ is being evaluated. If $a^j$ corresponds with a specific entry in $NL$, then it is either $TL$ or $FL$.
\begin{equation}
    NL_{ij} = \begin{cases} 
     TL, & D_{i} \xrightarrow{} NL_{j}\\
     FL, & D_{i} \not\to NL_{j}
    \end{cases} 
    \label{eq:NoiseLabel Logic}
\end{equation}

Weights and biases are also three-dimensional, so each dataset–label combination has its own parameters. With single-neuron ANNs, weights are $W_{N_D^2 \times c \times 1}$ and biases $B_{N_D^2 \times 1 \times 1}$, initialized via Xavier. These matrices are re-initialized each iteration and are not trainable outside of the SDTI layer. (Alg. \ref{alg:sdti_workflow}, Line \ref{line:winit}).

\subsection{Forward Propagation}
Matrix multiplication of two three-dimensional matrices with matching 0-th dimensions produces a dot product for each corresponding 2D slice. Each output corresponds to a unique dataset–target combination, using its specific weight and bias. This weighted sum is passed through a Sigmoid function $\sigma$, yielding the SDTI layer output $\hat{O}$ (Alg. \ref{alg:sdti_workflow}, Line \ref{line:forwardprop}, Eq. \ref{eq:MatMul}).
\begin{equation}
    \hat{O}_{N_D^2 \times m \times 1} = \sigma((A_{N_D^2 \times m \times c} \cdot W_{N_D^2 \times c \times 1})+B_{N_D^2 \times 1 \times 1}) 
    \label{eq:MatMul}
\end{equation}
\subsection{Updating Weights}
This output ($\hat{O}$) is then passed into a binary cross-entropy loss (BCE) function that is capable of calculating the losses for all $N_D^2$ data and target variable combinations simultaneously (Alg. \ref{alg:sdti_workflow} Line. \ref{line:cost}). These costs are then used to inform an Adam optimizer \cite{b17} together with $H$ to update all weights and biases simultaneously while still being independent of each other (Alg. \ref{alg:sdti_workflow} Line \ref{line:update}).

\subsection{Predictions}

Once the SDTI layer within the inner loop is completed, the costs from the final SDTI iteration are reshaped from \ref{eq:cost_output}
\begin{equation}
    \hat{O} \in \mathbb{R}^{N_D^2 \times 1} \xrightarrow{} \hat{O^2} \in \mathbb{R}^{N_D \times N_D}
    \label{eq:cost_output}
\end{equation}
\noindent where each row corresponds to a dataset and each column corresponds to a target variable. This output ($\hat{O^2}$) is then multiplied by $-1$ and the maximum value for each row is identified as seen in \ref{eq:argmax_cost_ouuput}:
\begin{equation}
    y_{i, N_D \times N_D} = \arg\max_j \left(-\hat{O}_{i,j}\right), 
    \quad i = 1, \dots, N_D
    \label{eq:argmax_cost_ouuput}
\end{equation}
The resulting predictions are appended to a data structure $T$ that tracks all predictions across training epochs (Alg. \ref{alg:sdti_workflow} Line. \ref{line:append}). After all iterations are completed, the predictions are aggregated by summing along the first axis and applying the argmax (Alg. \ref{alg:sdti_workflow} Line. \ref{line:prediction}) as seen below \ref{eq:final_output}:
\begin{equation}
    \hat{y}_{i, {N_D}} = \arg\max_j \left( \sum_{k=1}^{E} T^{(k)}_{i,j} \right), 
    \quad i = 1, \dots, N_D
    \label{eq:final_output}
\end{equation}

\section{Algorithm Structures}
\subsection{Manifold Complexity}
A manifold, in this context, represents the space of all possible parameter values for a model’s weights and biases and is a lower dimensional representation of the relationship seen in the data and target variables\cite{b9}. For a model to minimize its cost, these parameters must traverse this space toward regions of lower error. As the semantic relationships within a dataset become more entangled or weakly defined, the corresponding manifold grows more complex and irregular.
The SDTI framework leverages this property by evaluating all possible dataset–target variable combinations, introducing controlled stochasticity. When data and target variables are mismatched, their semantic relationship is partially corrupted, distorting the underlying function and yielding a manifold that is more chaotic and difficult to optimize.

The SDTI layer does not aim for each ANN to learn deployable representations but to probe the geometry of its corresponding manifold. The cost trajectory over iterations serves as an indicator of underlying geometric complexity. The final cost thus reflects the manifold’s structural difficulty, and the model identifies the target variable producing the lowest cost as the most semantically aligned pairing.
Mismatched pairs produced manifolds far more resistant to optimization than anticipated, creating a distinct signal separating correct from incorrect pairings (Table \ref{tab:combined_final}). This resistance revealed that manifold complexity itself functions as an implicit supervisory signal—enabling single-neuron ANNs to distinguish correct mappings without pretraining or deep architectures, consistent with findings that representation geometry can guide learning even in the absence of labels \cite{b22,b23}.

\begin{table}[htbp]
\centering
\footnotesize
\begin{tabular}{|c|c|c|}
\hline
\textbf{Group} & \textbf{Mean Cost ($\mu$)} & \textbf{Variance ($\sigma^2$)} \\
\hline
Correct Pairs & 0.6483 & 0.2597 \\
Incorrect Pairs & 0.9913 & 0.6093 \\
\hline
\end{tabular}
\caption{Mean and variance of cost values for correct vs. incorrect dataset–label pairings for each individual run in abalation study with over 30 SDTI epochs.}
\label{tab:variance_summary}
\end{table}

\begin{figure}[htbp]
    \centering
    \includegraphics[width=1\linewidth]{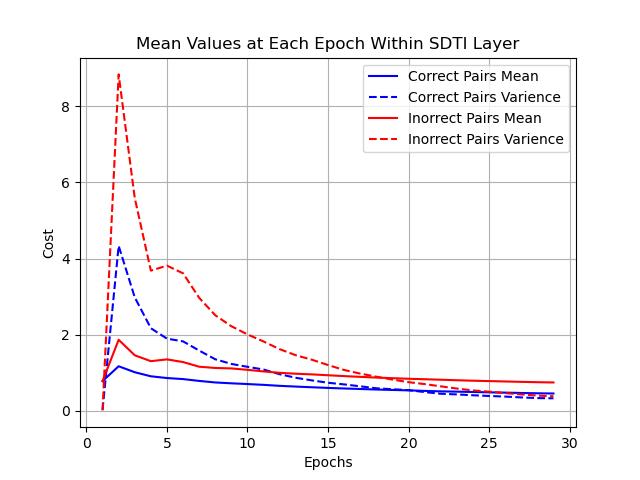}
    \caption{Stratified representation of mean values and variances for every SDTI epoch in all ablation study results that included 30 SDTI epochs.}
    \label{fig:ManifoldVis}
\end{figure}

We compared cost variances $\sigma^2_{\text{correct}}$ and $\sigma^2_{\text{incorrect}}$ across 30 SDTI iterations to test whether incorrect dataset–target variable pairings showed greater instability. Using Levene’s and Brown–Forsythe tests ($p = 8.77\times10^{-195}$ and $p = 1.77\times10^{-54}$), we rejected $H_0: \sigma^2_{\text{correct}} = \sigma^2_{\text{incorrect}}$ in favor of $H_a: \sigma^2_{\text{correct}} < \sigma^2_{\text{incorrect}}$. The results confirm substantially higher cost variance for incorrect pairings, suggesting more complex and irregular manifolds in their underlying loss landscapes.

Each dataset–target variable combination $(D_i, Y_j)$ defines a data manifold:
\begin{equation}
    \mathcal{M}_{ij} = \{(x, y_j) \mid x \in D_i\}.
\end{equation}

When labels are mismatched ($Y_j \neq Y_i$), this manifold becomes geometrically distorted, resulting in higher curvature and irregularity in the corresponding loss surface.
The optimizer experiences this as increased gradient variance:

\begin{equation}
\mathrm{Var}\!\left[\nabla_{\theta} \mathcal{L}_{ij}\right] 
> 
\mathrm{Var}\!\left[\nabla_{\theta} \mathcal{L}_{ii}\right].
\end{equation}
which is consistent with the empirical results in Table~\ref{tab:variance_summary} and Figure~\ref{fig:ManifoldVis}. 
Thus, the expected loss itself acts as a proxy for manifold smoothness:
The correct dataset–target variable pairing yields the most stable and least complex manifold.

In essence, SDTI exploits the fact that incorrect data–target variable pairings yield higher loss and greater optimization variance. 
By comparing losses across all combinations in parallel, the framework identifies the pairing that converges most smoothly—corresponding to the correct task.
This theoretical property explains why SDTI can successfully infer the correct target variable in a zero-shot setting without pretraining or external supervision.

\section{Results and Discussion}
After running the ablation study across a total of 3600 parameter combinations, the following Figure \ref{fig:ablation_results} summarizes the results.

\begin{figure}[htbp]
    \centering
    \includegraphics[width=1\linewidth]{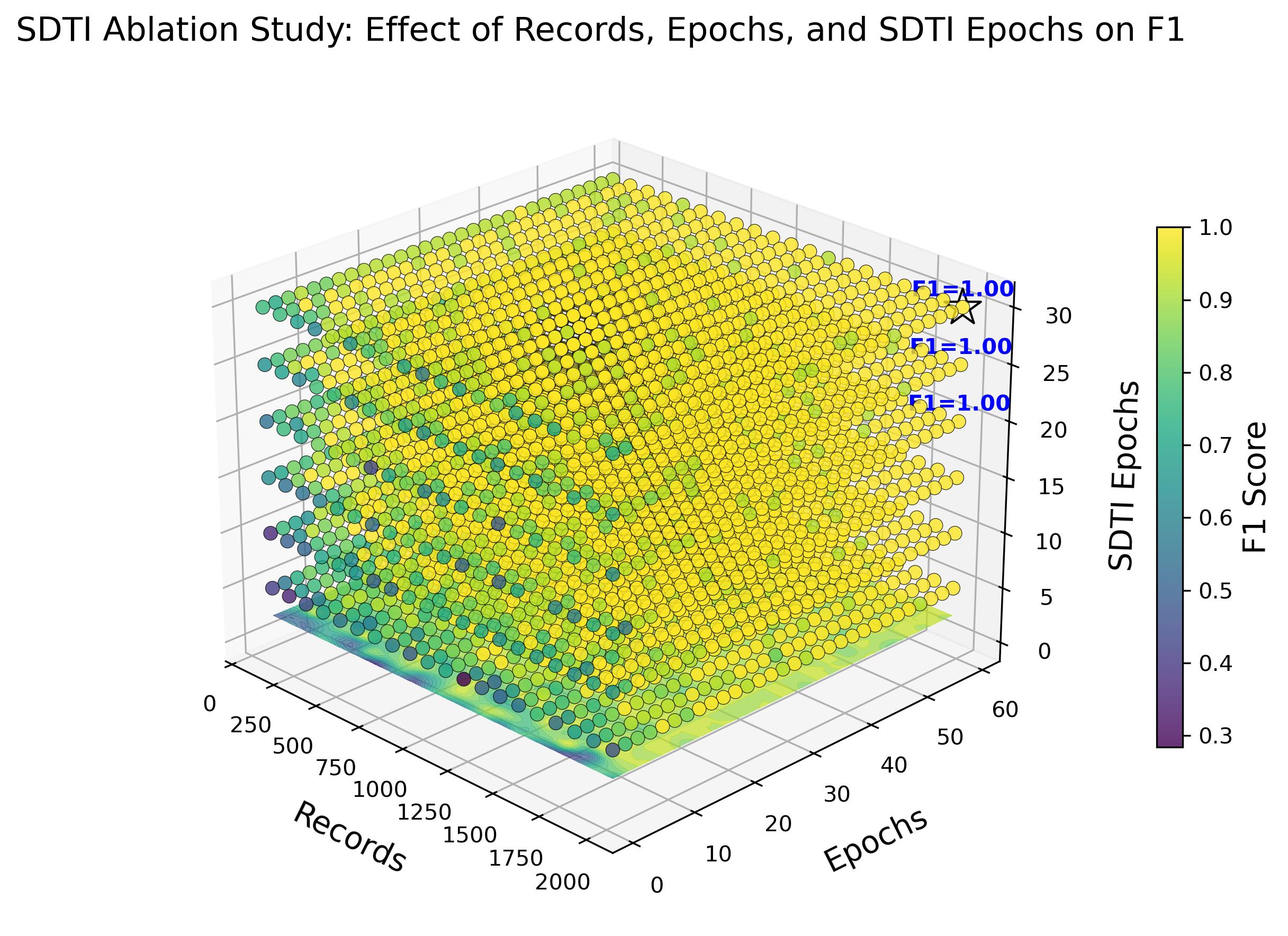}
    \caption{Ablation study over SDTI hyperparameters. Each point represents one configuration in the (A,B,C) parameter space, where A is the number of training records, B is the number of epochs, and C is the number of SDTI iterations. Color intensity encodes the resulting F1 score. Regions with more than 500 records and 10+ epochs achieve near-perfect alignment (F1 $>$ 0.98), confirming the stability of SDTI under sufficient data and training time.}
    \label{fig:ablation_results}
\end{figure}

\begin{table*}[htbp]
\centering
\footnotesize
\resizebox{\textwidth}{!}{%
\begin{tabular}{|c|cc|cc|cc|cc|}
\hline
\textbf{Field} 
& \multicolumn{2}{c|}{\textbf{Records ($\geq$500 vs $<$500)}} 
& \multicolumn{2}{c|}{\textbf{Epochs ($\geq$10 vs $<$10)}} 
& \multicolumn{2}{c|}{\textbf{SDTI-Epochs (30 vs 5)}} 
& \multicolumn{2}{c|}{\textbf{Parameters (Improved vs Worse)}} \\
\cline{2-9}
& $\geq$ 500 . & $<$ 500 & $\geq$ 10 & $<$ 10 & 30 & 5 & Imp. & Worse. \\
\hline
Count & 2880 & 720 & 3000 & 600 & 600 & 600 & 400 & 20 \\
Mean F1& 0.9705 & 0.9352 & 0.9830 & 0.8656 & 0.9724 & 0.9315 & 0.9900 & 0.7193 \\
F1 STD & 0.0733 & 0.0963 & 0.0377 & 0.1398 & 0.0597 & 0.1095 & 0.0262 & 0.1712 \\
Min F1& 0.2843 & 0.3529 & 0.6863 & 0.2843 & 0.5686 & 0.2843 & 0.9216 & 0.3529 \\
25th Percentile F1& 1.0000 & 0.9216 & 1.0000 & 0.7745 & 0.9216 & 0.7745 & 1.0000 & 0.6358 \\
50th Percentile F1& 1.0000 & 1.0000 & 1.0000 & 0.9216 & 1.0000 & 0.9216 & 1.0000 & 0.7647 \\
75th Percentile F1& 1.0000 & 1.0000 & 1.0000 & 1.0000 & 1.0000 & 1.0000 & 1.0000 & 0.8431 \\
Max F1& 1.0000 & 1.0000 & 1.0000 & 1.0000 & 1.0000 & 1.0000 & 1.0000 & 0.9216 \\
\hline
\end{tabular}%
}
\caption{Comprehensive comparison across all parameters in abalation study. Columns correspond to the original comparisons: records, epochs, SDTI-epochs, and parameter settings. For each pair, the first column represents the higher/better condition (improved parameters: records $\geq $500, epochs $\geq$ 10, 30 SDTI-epochs, worse parameters: records $<$ 500, epochs $<$ 10, 5 SDTI-epochs) and the second column represents the lower/worse condition. In each sub-table, only the indicated parameter is varied, while all other parameters are held constant.}
\label{tab:combined_final}
\end{table*}

The results in Table~\ref{tab:combined_final} show a measurable difference in performance based on the number of records used. Runs with more than 500 records achieved a higher mean F1 score (0.97 vs. 0.94) and lower mean standard deviation (0.07 vs. 0.09), indicating improved stability. In addition, these runs reached perfect F1 scores as early as the 25th percentile, while smaller-record runs required higher percentiles to reach this threshold. Despite the stronger performance of larger datasets, the maximum number of records used was only 2000, suggesting that relatively modest sample sizes can sufficiently capture the underlying structure in a single-neuron ANN. This is advantageous for efficiency in a production environment.

Table~\ref{tab:combined_final} also highlights the strong effect of training epochs. Configurations with at least 10 epochs significantly outperformed those with fewer epochs, both in terms of mean F1 (0.98 vs. 0.87) and a lower mean standard deviation (0.04 vs. 0.14). Moreover, the ability to achieve an F1 score of 1 appeared much earlier in the distribution (25th percentile vs. 75th percentile). These results suggest that a sufficient number of epochs is critical for optimization in the SDTI framework.

In contrast, Table~\ref{tab:combined_final} also shows that varying the number of SDTI-epochs had less impact on performance compared to training epochs. Both groups achieved identical maxima, and the difference in mean F1 was modest (0.97 vs. 0.93). The main benefit of higher SDTI-epochs was reduced mean standard deviation (0.06 vs. 0.11), suggesting that additional iterations improved stability rather than raw predictive performance. This result indicates that SDTI performance is more sensitive to training epochs and record count than to the number of SDTI epochs.

The results in Table~\ref{tab:combined_final} show a clear performance gap between models trained under stronger parameter settings (\(\geq 500\) records, \(\geq 10\) epochs, 30 SDTI-epochs) and those trained under weaker settings (\(< 500\) records, \(< 10\) epochs, \(\leq 5\) SDTI-epochs). Across 400 runs in the improved regime, the mean F1 score was 0.99 with very low standard deviation (0.026), and the majority of runs (\(\geq 25th\) percentiles) achieved a perfect F1 of 1.0. Furthermore, the stronger parameters produced a mean time to completion of 98 seconds with a standard deviation of 55 seconds. In contrast, the 20 runs under the worse parameters had a mean F1 of only 0.72, with substantially higher variability (standard deviation = 0.171). Their lower quartile values show that at least 25\% of runs fell below 0.64, and no runs exceeded 0.92. 

This demonstrates that the chosen criteria---sufficient training data, adequate training epochs, and extended SDTI-epochs---are not only individually important but also collectively necessary for reliable convergence and high performance. The nearly perfect F1 scores in the improved condition further highlight the stability and robustness of the model when trained under these settings.

\begin{table}[htbp]
\centering
\footnotesize
\resizebox{\columnwidth}{!}{%
\begin{tabular}{|c|c|c|}
\hline
\textbf{Algorithm} & \textbf{Mean F1} & \textbf{Standard Deviation} \\
\hline
SDTI & 0.9900 & .0262 \\
Pearson's Correlation & 0.8431 & 0 \\
Mutual Information & 0.8431 & 0 \\
Cosine Similarity & 0.8431 & 0 \\
\hline
\end{tabular}%
}
\caption{Comparison of SDTI with benchmark algorithms.}
\label{tab:benchmark}
\end{table}

The results in Table \ref{tab:benchmark} demonstrate the superior performance of the SDTI model relative to traditional correlation-based benchmarks. Using the mean F1 score reported in Table \ref{tab:combined_final}, SDTI achieves a 14.69\% improvement over all baseline algorithms. Although SDTI exhibits a slightly higher standard deviation, the variance remains small relative to its mean performance, indicating consistent behavior across runs. The identical mean F1 scores and zero variance observed in the benchmark algorithms stem from their deterministic nature—these methods lack stochastic components or tunable parameters, producing identical predictions across iterations. While the continuous similarity values generated by Pearson’s Correlation, Mutual Information, and Cosine Similarity differed numerically, their discrete predictions after the \textit{argmax} operation were equivalent, leading to identical classification metrics. This shows deterministic limitations of correlation-based methods.

\section{Potential Applications}
The SDTI model has many wide-ranging applications. As mentioned above, a large focus of the AutoML community is automating different elements of the machine learning pipeline to decrease the requisite mathematical understanding of the researcher, with the ultimate goal of making machine learning much more accessible. The SDTI model is designed to automate the training pipeline a step further by being able to determine the correct target variable for a given dataset without being pre-trained on a large corpus of data. This would allow for the extremely time and labor-intensive task of curating a training pipeline for complicated tasks, such as an LLM for users to interact with. With the support of the SDTI model and similar modules, the LLM could become a central component in a virtual assistant application that can continue to source, learn, and synthesize data much faster without needing a training pipeline heavily curated by humans. 

Additionally, hyperparameter tuning often requires large batches of models to test different combinations of hyperparameters and how they affect model performance. There are many methodologies with the ultimate goal of providing the optimal hyperparameters with the smallest number of trained models. A vectorized implementation of model architectures similar to the SDTI layer could allow for multiple sets of hyperparameters to be tested at once on a sample of data if said data is reflective of the entire dataset. This methodology is liable to run into memory constraints, and a batching mechanism is necessary to implement in a production environment. 

The vectorized implementation of models seen in the SDTI layer could also be leveraged for meta-learning by treating each model as an agent in a specific task. Much like hyperparameter tuning, the vectorized implementation of model methodology could decrease the overall training time required, as multiple agents could be run simultaneously. 

\section{Conclusion}
This paper presents both theory and experiments in self-directed task identification (SDTI), leveraging the strengths of the SDTI model for zero-shot learning. A central insight is that the manifold of all possible trainable parameter values in the SDTI layer depends not only on the dataset itself but also on the target variable. When an ANN is trained on a dataset with an incorrect target variable, the resulting manifold becomes more complex, which in turn makes optimization more difficult. SDTI exploits this effect by comparing performance across ANNs trained with every data-target combination; the model associated with the correct target variables can be identified because it converges more efficiently and achieves a lower cost. This mechanism enables SDTI to predict the correct target variable for a dataset from among a pool of noisy candidates, thereby validating the methodology at the foundational and proof-of-concept level.  

Even with the only hyperparameter optimization being random sampling, the distinction remains clear: for a given dataset, an ANN trained with the correct target variable consistently yields a lower cost than one trained with an incorrect target, precisely due to the difference in manifold complexity. In the context of the SDTI model, this relationship suggests that manifold complexity not only highlights mismatched targets but also provides a natural inductive bias that simplifies zero-shot prediction. By tracking the cost gap between correct and incorrect target assignments, the SDTI framework gains robustness in identifying ground truth without additional supervision and increased computational efficiency. Future work will address key limitations. The primary objective is making modifications to the SDTI model allowing it to work with imbalanced datasets and multimodal data so that it could be more properly integrated production level automated data annotation pipelines in a variety of settings.
{
    \small
    \bibliographystyle{elsarticle-num}
    \bibliography{ref}

@article{b1,
  author    = {Wenwu Zhu and Xin Wang and Pengtao Xie},
  title     = {Semi-autonomous machine learning},
  journal   = {AI Open},
  volume    = {3},
  pages     = {58-70},
  year      = {2022},
  issn      = {2666-6510},
  doi       = {10.1016/j.aiopen.2022.06.001},
  url       = {https://doi.org/10.1016/j.aiopen.2022.06.001}
}

@article{alom2019state,
  title={A state-of-the-art survey on deep learning theory and architectures},
  author={Alom, Md Zahangir and Taha, Tarek M and Yakopcic, Chris and Westberg, Stefan and Sidike, Paheding and Nasrin, Mst Shamima and Hasan, Mahmudul and Van Essen, Brian C and Awwal, Abdul AS and Asari, Vijayan K},
  journal={electronics},
  volume={8},
  number={3},
  pages={292},
  year={2019},
  publisher={Multidisciplinary Digital Publishing Institute}
}

@article{b2,
  author    = {Quanming Yao and Mengshuo Wang and Yuqiang Chen and Wenyuan Dai and Yu-Feng Li and Wei-Wei Tu and Qiang Yang and Yang Yu},
  title     = {Taking human out of learning applications: A survey on automated machine learning},
  journal   = {arXiv preprint arXiv:1810.13306},
  year      = {2018},
  url       = {https://arxiv.org/abs/1810.13306}
}

@article{b3,
  author    = {Marc-André Zöller and Marco F. Huber},
  title     = {Benchmark and survey of automated machine learning frameworks},
  journal   = {Journal of artificial intelligence research},
  volume    = {70},
  pages     = {409-472},
  year      = {2021}
}

@article{b4,
  author    = {Colin White and Mahmoud Safari and Rhea Sukthanker and Binxin Ru and Thomas Elsken and Arber Zela and Debadeepta Dey and Frank Hutter},
  title     = {Neural architecture search: Insights from 1000 papers},
  journal   = {arXiv preprint arXiv:2301.08727},
  year      = {2023},
  url       = {https://arxiv.org/abs/2301.08727}
}

@inproceedings{b5,
  author    = {Adam Santoro and Sergey Bartunov and Matthew Botvinick and Daan Wierstra and Timothy Lillicrap},
  title     = {Meta-learning with memory-augmented neural networks},
  booktitle = {International conference on machine learning},
  pages     = {1842-1850},
  year      = {2016},
  organization = {PMLR}
}

@inproceedings{b6,
  author    = {Ronghang Hu and Amanpreet Singh},
  title     = {Unit: Multimodal multitask learning with a unified transformer},
  booktitle = {Proceedings of the IEEE/CVF International Conference on Computer Vision},
  pages     = {1439-1449},
  year      = {2021}
}

@incollection{b7,
  author    = {Yann Lecun},
  title     = {Learning processes in an asymmetric threshold network},
  booktitle = {Disordered systems and biological organization},
  editor    = {E. Bienenstock and F. Fogelman-Soulie and G. Weisbuch},
  pages     = {233-240},
  year      = {1986},
  publisher = {Springer-Verlag},
  address   = {Les Houches, France}
}

@article{b9,
  author    = {German Magai and Anton Ayzenberg},
  title     = {Topology and geometry of data manifold in deep learning},
  journal   = {arXiv preprint arXiv:2204.08624},
  year      = {2022},
  url       = {https://arxiv.org/abs/2204.08624}
}

@article{b16,
  author    = {Belkin, Mikhail and Niyogi, Partha and Sindhwani, Vikas},
  title     = {Manifold regularization: A geometric framework for learning from labeled and unlabeled examples},
  journal   = {Journal of Machine Learning Research},
  volume    = {7},
  number    = {11},
  pages     = {2399--2434},
  year      = {2006}
}

@inproceedings{b17,
  author    = {Kingma, Diederik P. and Ba, Jimmy},
  title     = {Adam: A Method for Stochastic Optimization},
  booktitle = {Proceedings of the 3rd International Conference on Learning Representations (ICLR)},
  year      = {2015},
  url       = {https://arxiv.org/abs/1412.6980}
}

@inproceedings{b18,
  author    = {Chen, Ting and Kornblith, Simon and Norouzi, Mohammad and Hinton, Geoffrey},
  title     = {A Simple Framework for Contrastive Learning of Visual Representations},
  booktitle = {Proceedings of the 37th International Conference on Machine Learning (ICML)},
  year      = {2020},
  url       = {https://arxiv.org/abs/2002.05709}
}

@inproceedings{b19,
  author    = {Grill, Jean-Bastien and Strub, Florian and Altché, Florent and Tallec, Corentin and Richemond, Pierre H. and Buchatskaya, Elena and Doersch, Carl and Avila Pires, Bernardo and Guo, Zhaohan Daniel and Gheshlaghi Azar, Mohammad and Piot, Bilal and Kavukcuoglu, Koray and Munos, Rémi and Valko, Michal},
  title     = {Bootstrap Your Own Latent: A New Approach to Self-Supervised Learning},
  booktitle = {Advances in Neural Information Processing Systems (NeurIPS)},
  year      = {2020},
  url       = {https://arxiv.org/abs/2006.07733}
}

@inproceedings{b20,
  author    = {Radford, Alec and Kim, Jong Wook and Hallacy, Chris and Ramesh, Aditya and Goh, Gabriel and Agarwal, Sandhini and Sastry, Girish and Askell, Amanda and Mishkin, Pamela and Clark, Jack and Krueger, Gretchen and Sutskever, Ilya},
  title     = {Learning Transferable Visual Models From Natural Language Supervision},
  booktitle = {International Conference on Machine Learning (ICML)},
  year      = {2021},
  url       = {https://arxiv.org/abs/2103.00020}
}

@inproceedings{b21,
  author    = {Zhou, Kaiyang and Yang, Jingkang and Loy, Chen Change and Liu, Ziwei},
  title     = {Learning to Prompt for Vision-Language Models},
  booktitle = {International Journal of Computer Vision (IJCV)},
  year      = {2022},
  url       = {https://link.springer.com/article/10.1007/s11263-022-01653-1}
}

@inproceedings{b22,
  author    = {Caron, Mathilde and Touvron, Hugo and Misra, Ishan and Jégou, Hervé and Mairal, Julien and Bojanowski, Piotr and Joulin, Armand},
  title     = {Emerging Properties in Self-Supervised Vision Transformers},
  booktitle = {Proceedings of the IEEE/CVF International Conference on Computer Vision (ICCV)},
  year      = {2021},
  url       = {https://openaccess.thecvf.com/content/ICCV2021/html/Caron_Emerging_Properties_in_Self-Supervised_Vision_Transformers_ICCV_2021_paper.html}
}

@inproceedings{b23,
  author    = {Rifai, Salah and Dauphin, Yann and Vincent, Pascal and Bengio, Yoshua and Muller, Xavier},
  title     = {The Manifold Tangent Classifier},
  booktitle = {Advances in Neural Information Processing Systems (NeurIPS)},
  year      = {2011},
  url       = {https://papers.nips.cc/paper/2011/hash/4409-the-manifold-tangent-classifier.html}
}
}


\end{document}